\pdfoutput=1
\documentclass[sigconf]{acmart}

\usepackage{comment}
\usepackage{booktabs} 
\usepackage{bm}
\usepackage[subrefformat=parens]{subcaption}
\usepackage{caption}

\setcopyright{none}

\acmDOI{}

\acmISBN{}

\acmConference[]{}{}{} 
\acmYear{}
\copyrightyear{}

\acmPrice{}

\begin{document}
\title{Cross-domain Recommendation via Deep Domain Adaptation}

\author{Heishiro Kanagawa}
\authornote{This work was conducted during the author's intership at Yahoo! JAPAN.}
\affiliation{%
    \institution{Yahoo Japan Corporation}
  \city{Tokyo} 
  \state{Japan} 
}
\email{heishiro.kanagawa@gmail.com}

\author{Hayato Kobayashi}
\affiliation{%
  \institution{Yahoo Japan Corporation}
  \city{Tokyo} 
  \state{Japan} 
}
\email{hakobaya@yahoo-corp.jp}

\author{Nobuyuki Shimizu}
\affiliation{%
  \institution{Yahoo Japan Corporation}
  \city{Tokyo} 
  \state{Japan} 
}
\email{nobushim@yahoo-corp.jp}

\author{Yukihiro Tagami}
\affiliation{%
  \institution{Yahoo Japan Corporation}
  \city{Tokyo} 
  \state{Japan} 
}
\email{yutagami@yahoo-corp.jp}
\author{Taiji Suzuki}
\affiliation{%
  \institution{The University of Tokyo}
  \city{Tokyo}
  \state{Japan}
}
  \email{taiji@mist.i.u-tokyo.ac.jp}


\begin{abstract}
The behavior of users in certain services could be a clue that can be used to infer their preferences and may be used to make recommendations for other services they have never used. 
However, the cross-domain relationships between items and user consumption patterns are not simple, especially when there are few or no common users and items across domains. 
To address this problem, we propose a content-based cross-domain recommendation method for cold-start users that does not require user- and item- overlap. 
We formulate recommendation as extreme multi-class classification where labels (items) corresponding to the users are predicted. 
With this formulation, the problem is reduced to a domain adaptation setting, 
in which a classifier trained in the source domain is adapted to the target domain. 
For this, we construct a neural network that combines an architecture for domain adaptation, Domain Separation Network, with a denoising autoencoder for item representation. 
We assess the performance of our approach in experiments on a pair of data sets collected from movie and news services of Yahoo! JAPAN
and show that our approach outperforms several baseline methods including a cross-domain collaborative filtering method.
\end{abstract}

%
%
\ccsdesc[300]{Information systems~Recommender systems}


\keywords{Recommender systems, Cross-domain recommendation,
Deep Learning, Domain Adaptation, Extreme Classification}

\maketitle

\section{Introduction}\label{sec:introduction}
Traditional recommender systems usually assume user's past interactions (e.g. ratings or clicks) to make meaningful recommendations.  
While this seems natural, however, the utility of such system becomes degraded when the assumption does not hold, 
e.g. when a new user arrives at the service or when we aim to perform cross-selling of products from an unused service. 
On the other hand, as the variety of web services has increased, information about cold-start users can be obtained from their activities in other services. 
Therefore, cross-domain recommender systems, which leverage such additional information from other related domains, have gained research attention in recent years 
as a promising solution to the user cold-start problem \cite{Cantador2015}. 

In this paper, we deal with cross-domain recommendations for cold-start users, particularly in the absence of common users and items. 
Although it is possible to suggest items and obtain feedback from the target users \cite{Rubens2015}, 
we aim to build a recommender system that can work even when such alternative is not available. 
In this situation, a major challenge is that 
traditional methods can not be used for the elicitation of the relations between the two services due to the lack of common users. 


A naive approach to overcome this difficulty would be profile-based recommendations 
in which domain-independent features, such as gender or occupation, are used to obtain user representations shared across domains. 
While this approach allows the application of single-domain methods, 
such features are often hard to obtain as users often hesitate to reveal their profiles due to privacy concerns. 
Therefore, without such features, we need to represent users by their consumption patterns, 
with which we can utilize the labeled data in a source domain 
to learn the relation between users and items.  

Previous studies dealt with the above challenge by using specific forms of auxiliary information
such as search queries \cite{Elkahky:2015}, 
social tags \cite{Abel2011, Abel2013}, 
and external knowledge repositories like Wikipedia \cite{soton66281, Fernandez-Tobias2011, Kaminskas2013, Kaminskas2014}, 
which may not be available in some applications. 
For instance, the information about a TV program that is only available in a particular service
may not be available on Wikipedia because few people have written about it. 
On the other hand, collaborative filtering (CF) approaches were also proposed \cite{Zhang:2010, Iwata:2015}. 
While CF methods do not require auxiliary information, they suffer from data sparsity and require 
a substantial amount of previous user interactions to give a high-quality recommendation. 

In contrast, this paper investigates content-based approaches. 
In particular, we study a deep-learning apparoach to cross-domain recommendation. 
Deep learning has been successfully applied to recommender systems\cite{Oord:2013, Wang:2015, Elkahky:2015, Covington:2016, Wu2016, Zheng2016, Wang2016, Wu2017, He2017}. 
In addition to the success in the applications to recommender system, 
deep learning methods show better performance in the area of transfer learning because of its ability to learn transferable features from the data. 
In fact, in the area of domain adaptation, deep neural networks show state-of-the-art performance in computer vision and natural language processing tasks \cite{Ganin:2016, NIPS2016_6254}. 
Based on this observation, we hypothesized that domain adaptation with deep neural networks could also be applied to cross-domain recommendation. 

Domain adaptation \cite{citeulike3744578} is a technique to learn a new domain (the target domain) with a few or no labeled training data using 
the knowledge acquired from another domain that has labeled data (the source domain). 
With domain adaptation, a classifier trained in the source domain can be applied to the target domain.  
Our approach is to treat recommendation as extreme multi-class classification \cite{Partalas2015, Covington:2016} where labels (items) corresponding to a user history are predicted. 
A benefit of this approach is that the problem of cross-domain recommendation can be seen as an instance of domain adaptation, which has been extensively studied in the machine learning literature. 
To achieve domain adaption, we use a newly proposed neural network architecture for unsupervised domain adaptation, 
domain separation network (DSN) proposed by \citet{NIPS2016_6254}. 
In addition, to reduce the difficulty of extreme classification and deal with new items, 
we combine item features via a Stacked Denoising Auto Encoder (SDAE) \cite{Vincent2010}. 
To validate our approach, we performed experiments on real-world datasets collected from the movie and news services of Yahoo! JAPAN. 
In summary, our contributions in this paper are as follows:
\begin{enumerate}
  \item We proposed a deep domain adaptation approach for cross-domain recommendation. 
  \item We examined the capacity of our method in experiments on large-scale real-world datasets and confirm the effectiveness of domain adaptation. 
  \item We also showed that our approach has better performance compared to a state-of-the-art cross-domain recommendation method. 
\end{enumerate}

The rest of this paper is organized as follows.
In Section \ref{sec:related_work}, we give a summary of related works. 
In Section \ref{sec:prelim}, we first formulate the problem of cross-domain recommendation and introduce the machine learning methods that we use for out model.  
In Section \ref{sec:proposal}, we elaborate our proposed model. 
In Section \ref{sec:experiment}, we describe the datasets we used for the study and evaluate our method. 
Finally, in Section \ref{sec:conclusion}, we conclude with a discussion on the future work.

\section{Related Work}\label{sec:related_work}
\subsection{Cross-Domain Recommendations}
The focus of our work is on addressing the user cold-start problem by leveraging user activities in other domains. 
One of the major challenges in cross-domain recommendation is the lack of labelled data
, which is essential to establish the link between domains.  
Approaches to this problem can be broadly classified into two types: collaborative filtering and content-based filtering. 

Collaborative filtering suggests items based on the rating patterns of users by taking into account the ones of other users with similar tastes. 
As the only required data is rating patterns, collaborative filtering is a generally applicable approach. 
By assuming the existence of shared items or users, single-domain methods, such as matrix factorization \cite{SalMnih08, Salakhutdinov:2008}, can be applied. 
However, in the absence of user/item overlaps, as no users in the target domain have rated items in the source domain, collaborative filtering does not work \cite{Cremonesi:2011}. 
Two approaches have been proposed to overcome this limitation \cite{Zhang:2010, Iwata:2015}. 
In particular, \citet{Iwata:2015} propose a method which is based on Bayesian probabilistic matrix factorization\citep{Salakhutdinov:2008}. 
Their method assumes latent vectors are generated from a common Gaussian distribution with a full covariance matrix, 
which makes latent factors aligned across domains and enables predictions in the target domain.  

In contrast, our approach is a content-based approach. 
Content-based filtering recommends items by comparing the content of the items with a user profile in accordance with what they consumed before. 
Unlike collaborative filtering, a domain link can be established without an item/user overlap 
by using the auxiliary information about users or items. 
\citet{Abel2011, Abel2013} presented an approach that constructs common user representations from social tags, such as on Flickr. 
\citet{Fernandez-Tobias2011} and \citet{Kaminskas2013} proposed a method for recommending music to place of interests. 
Their method is a know-ledge based framework built on the DBpedia ontology, which is used to measure the relatedness between the items in different domains. 
In their method, items need to be defined on the knowledge repository. 
Related to these studies, \citet{soton66281} proposed a method that uses Wikipedia as an universal vocabulary to describe items. Their method is to build a graph whose nodes correspond to Wikipedia articles describing items rated by the users, and edges express the relationships between those articles. 
A Markov chain model is defined using the graph and produces recommendation by assessing the probability of traversing from the nodes expressing a user to particular an item.
When articles that directly correspond to items do not exist, they proposed using tags or words associated with items to search for related Wikipedia articles. 
This requires extra preprocessing of the data, and descriptive articles may not exist.  
On the otherhand, our method has broader applicability as it only assumes the content information of items. 
\subsection{Deep Learning and Recommender Systems}
Deep learning is a powerful techqnique to learn predictive features from data. 
The representation of the user or item is essential in recommendation tasks
as the performance of recommendation is dependent on the ability to describe user preference and complex user-item relations.
For this reason, the application of deep learning to recommender systems has gained research attention. 
\citet{Wang:2015} and \citet{Li2015} combined autoencoder models with collaborative filtering; 
\citet{Wang2016} used a RNN model for item representation and combined it with CF; \citet{Wu2017} modeled temporal dynamics between users and items wirh RNNs. 

Deep neural networks are also used for content-based filtering: music recommendation \citep{Oord:2013}, news recommendation \citep{KJOh:2014}, and movie recommendation \citep{Covington:2016}.
Among the previous works, the most relevant to our setting is \citet{Elkahky:2015}, which dealt with user modeling for cross-domain recommendations using deep neural networks. 
They proposed the multi-view deep structured semantic model (MV-DSSM), which maps users and items into a latent space where the similarity between them is maximized. 
Their model requires the user's search queries or user overlaps to obtain user representations shared across domains. 
Our model, in contrast, does not require those types of data as our model obtain shared user representations via unsupervised domain adaptation. 
Our model can be seen as an extension of \citet{Covington:2016} and \citet{Wang:2015}. 
\citet{Covington:2016} proposed a collaborative filtering approach via extreme classification in which a classifier is trained to 
generate a candidate set. 
Inspired by their work, we approach to cross-domain recommendation with a content-based approach and domain adaptation so that 
different domains can be linked with semantic information. 
Also, as in \citet{Wang:2015}, to perform predictions on items that are not frequently observed, we combine item information with the model using a denoising autoencoder.

\section{Preliminaries}\label{sec:prelim}
\subsection{Problem setting}
In this paper, we assume that the input data takes the form of implicit feedback such as click logs. 
We also assume access to the content information of items.  
Now we formally define our problem setting. 
Let $Y = \{1,\dots,L\}$ be a collection of items in a source domain that we wish to recommend.  
From user logs on these items, we have a labeled data set $\mathbf{X}_S = \{\mathbf{x}_i^{S}, y_i^{S}\}_{i=1}^{N_s}$.  
Here $y_i^S \in Y$ denotes a label that represents an item consumed by a user, 
and $\mathbf{x}_i^S$ is raw features representing a list of items used by the user before the item $y_i^S$. 
In a target domain, we only have an unlabeled data set $\{\mathbf{x}_i^{T}\}_{i=1}^{N_T}$ due to the lack of common users between the two domains. 
As with the source domain, $\mathbf{x}_i^{T}$ denotes a use history composed of items in the target domain. 
Our objective is to recommend items in the source domain to users in the target domain.  
More formally, by exploiting the labeled data $X_S$ and the unlabeled data $X_T$, we aim to construct a classifier $\eta(\mathbf{x}^T)$, which defines a probability distribution over items $Y$ given a user history $\mathbf{x}^T$ in the target domain. We model this classifier with domain adaptation, which we explain in Section \ref{sec:da}. 

As an illustration, suppose that our task is to recommend videos to users who use a news browsing service. 
We have user logs of the video service in the source domain and of the news service in the target domain. 
A video can have text information about itself, such as descriptions of cast or plot. 
With this content information, we can construct raw user representations from a list of movies using the Bag of Words or the TF-IDF scheme. 
As with videos, we can construct representations for news users using textual information of news articles.  
Using content information, we obtain a labeled dataset for the source domain (the video service) and an unlabeled for the target domain (the news service). 
In this case, the task is to build a classifier that takes as input a news browsing history and assign probability to videos. 

\subsection{Domain Adaptation} \label{sec:da}
The labled data in the source domain would be useful in mining the relations between items and users; therefore, we want to utilize this data to learn a classifier for the target domain. 
Domain adaptation serves this purpose by {\sl adapting} a classifier trained on the labeled dataset in the source domain to the target domain. 
In the following, we formally define domain adaptation and introduce the architecture for domain adaptation. 
\subsubsection{Definition of Unsupervised Domain Adaptation}
Following \citep{Ganin:2016, NIPS2016_6254}, we define the problem of unsupervised domain adaptation. 
We consider a classification task where $X$ is the input space and $Y = \{1\dots, L\}$ is the label set.
Let $D_S$ and $D_T$ be probability distributions over $X\times Y$, which are assumed to be similar but different. 
We assume that we have an i.i.d sample of labeled data 
$\mathbf{X}_S = \{(\mathbf{x}_i^S, y_i^S)\}_{i=1}^{N_S}$
drawn from the source domain. 
On the other hand, we assume that little or no labeled data is available in the target domain. 
Instead, we have a large size of unlabeled i.i.d. sample 
$\mathbf{X}_T = \{\mathbf{x}_i^T\}_{i=N_S+1}^{N_S+N_T} \sim D_T^X$
, where $D_T^X$ is the marginal distribution over $X$ in the domain $T$. 
The goal of unsupervised domain adaptation is to obtain a classifier $\eta$ with a low {\sl target} risk defined by 
\begin{align*}
  R_T(\eta) = \mathrm{Pr}_{(\mathbf{x}, y) \sim D_T}\bigl(\eta(\mathbf{x}) \neq y\bigr)
\end{align*}
using the labeled data $\mathbf{X}_S$ and the unlabeled data $\mathbf{X}_T$. 

\subsubsection{Domain Separation Networks}
In this section, we introduce the neural network architecture we use in our method. 
We use domain separation networks (DSNs) \cite{NIPS2016_6254} to achieve domain adaptation. 
\begin{figure}[h]
  \centering
  \includegraphics[scale=0.48]{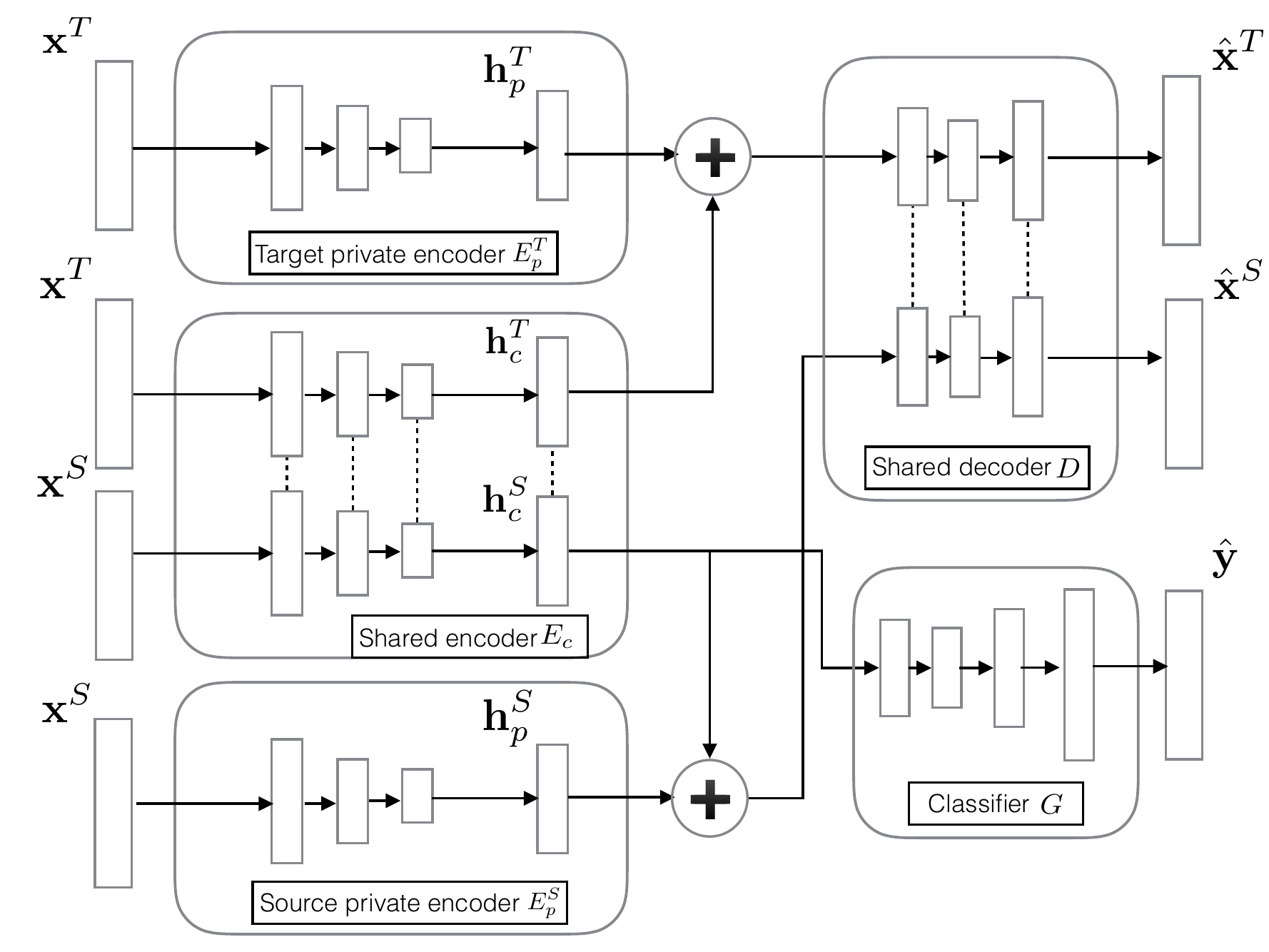}
  \caption{Architecture of a domain separation network \cite{NIPS2016_6254}}
  \label{fig:dann}
\end{figure}

As illustrated in Figure \ref{fig:dann}, a DSN model consists of the following components: 
a shared encoder $E_c(\mathbf{x}; \mathbf{\theta}_c)$, 
private encoders $E_p^S(\mathbf{x}; \mathbf{\theta}_p^S)$,  $E_p^T(\mathbf{x}; \mathbf{\theta}_p^T)$,
a shared decoder $D(\mathbf{h}; \mathbf{\theta}_d)$, 
and a classifier $G(\mathbf{h}; \mathbf{\theta}_g)$. 
Here $\theta_c$, $\theta_p^S$, $\theta_p^T$, $\theta_d$, and $\theta_g$ denote parameters for the shared encoder, the private encoder for the source and the target domain, the shared decoder, and the classifier, respectively. 

Given an input vector $\mathbf{x}$, which comes from the source domain or the target domain, a shared encoder function 
maps it to a hidden representation $\mathbf{h}_c$, which represents features {\sl shared} across domains. 
For the source (or target) domain, a DSN has a private encoder $E_p^S$ ($E_p^T$) that maps an input vector to 
a hidden representation $\mathbf{h}_p^S$ ($\mathbf{h}_p^T$), which serve as features that are specific to each domain. 
A decoder $D$ reconstructs an input vector $\mathbf{x}$ of the source (or target) domain from the sum of shared and private hidden representations $\mathbf{h}_c$, $\mathbf{h}_p^S$ ($\mathbf{h}_p^T$). 
We denote the reconstructed vector by $\hat{\mathbf{x}}$. 
The classifier $G$ takes a shared hidden representation $\mathbf{h}_c$ as input and predicts the corresponding labels. 

The training of this architecture is performed by minimizing the following objective function with respect to parameters 
$\mathbf{\theta}_c$, $\mathbf{\theta}_p^S$, $\mathbf{\theta}_p^T$, $\mathbf{\theta}_d$, $\mathbf{\theta}_g$: 
\begin{align}
  L_{\mathrm{DSN}} = L_{\mathrm{task}} + \alpha L_{\mathrm{recon}} + \beta L_{\mathrm{difference}} + \gamma L_{\mathrm{similarity}}.
  \label{eq:DSN_loss}
\end{align}
where $\alpha$, $\beta$, and $\gamma$ are parameters that control the effect of the loss terms. 

Here, $L_{\mathrm{task}}$ is a classification loss by which we train the model to predict the output labels. 
In the following, we  assume that the classification loss is the cross-entropy loss:
\begin{align*}
  L_{\mathrm{task}} = -\sum_{i=1}^{N_S} \mathbf{y}_i \cdot \log(\hat{\mathbf{y}}_i),
\end{align*}
where $\mathbf{y}_i$ is the one-hot encoding of the label of the $i$th example, 
and $\hat{\mathbf{y}}_i = G(E_c(\mathbf{x}_i^s))$ is the prediction
on the $i$-th input vector.  

$L_{\mathrm{recon}}$ is the reconstruction error defined as:
\begin{align*}
  L_{\mathrm{recon}} = \sum_{i=1}^{N_S} \|\mathbf{x}^S_{i} - \hat{\mathbf{x}}^S_i\|^{2} + 
  \sum_{i=N_S+1}^{N_S + N_T} \|\mathbf{x}^T_{i} - \hat{\mathbf{x}}^T_i\|^{2}.
\end{align*}
$L_{\mathrm{difference}}$ encourages the shared and private encoders to extract different type of features from the inputs.
As in \cite{NIPS2016_6254}, we impose a soft subspace orthogonality condition: 
\begin{align*}
  L_{\mathrm{difference}} = \left\|\mathbf{H}_c^S \bigl(\mathbf{H}_p^{S}\bigr)^{\top} \right\|_F^2 + \left\|\mathbf{H}_c^T \bigl(\mathbf{H}_p^{T}\bigr)^{\top}\right\|_F^2, 
\end{align*}
where $\|\cdot\|_F$ is the Frobenius norm. 
Here $\mathbf{H}_c^S$ is a matrix whose $i$-th row is the shared representation $\mathbf{h}_{c,i}^S = E_c(\mathbf{x}_i^S)$ of the $i$-th input vector in a source sample
, and $\mathbf{H}_p^S$ is similarly defined for private hidden representations. 
Likewise, $\mathbf{H}_c^T$ and $\mathbf{H}_p^T$ are defined for the target domain. 

Finally, the similarity loss $L_{\mathrm{similarity}}$ encourages the shared encoder to produce representations that are hardly distinguishable. 
For this, we use the domain adversarial similarity loss \cite{DBLP:conf/icml/GaninL15, Ganin:2016}. 
Let $d_i$ be a binary label that represents the domain of the $i$-th sample ($0$ denotes the source domain). 
To use the adversarial loss, we equip the DSN with a binary classifier $Z(\mathbf{h}_c; \mathbf{\theta}_z)$ parametrized by $\theta_z$, which predicts the domain labels of shared representations, and define the similarity loss as: 
\begin{align}
  L_{\mathrm{similarity}}
  = \sum_{i=1}^{N_S+N_T}\left\{ d_i \log{\hat{d}_i} + (1-d_i)\log(1-\hat{d}_i) \right\},
  \label{eq:DANN_loss}
\end{align}
where $\hat{d}_i$ is the predicted domain label of the $i$-th example. 
During training, the binary classifier and the shared encoder are trained adversarially; the loss $L_{\mathrm{similarity}}$ is minimized with respect to $\theta_z$ and maximized with respect to a parameter $\theta_c$ for the shared encoder $E_c$. 
The adversarial training can simply be performed by inserting the Gradient Reversal Layer (GRL) between the shared encoder and the domain label predictor. 
The GRL is a layer that acts as an identity function during the forward propagation, and during the back propagation, it flips the sign of propagated gradients. 
With GRL, we only have to minimize Eq. \eqref{eq:DANN_loss}. 

\subsection{Stacked Denoising Autoencoder}
We use a stacked denoising autoencoder (SDAE). 
It is a feed-forward neural network that learns a robust representation of the input data. 
Let $\mathbf{X}_I = \{\mathbf{x}_i\}_{i=1}^n$ be a set of input vectors  
and $\tilde{\mathbf{X}}_I = \{\tilde{\mathbf{x}}_i\}_{i=1}^n$ be a set of vectors each of which is corrupted with noise. 
The SDAE model takes a corrupted vector as input and then transform it to $\mathbf{h}$ with its encoder $E_{\theta}$.
The original input vector is predicted by reconstructing the hidden representation 
with its decoder $D_{\theta'}$. 
The network is trained to minimize the reconstruction error with respect to parameters $\theta$, $\theta'$:
\begin{align*}
  L_{\mathrm{SDAE}} = \sum_{i=1}^n \|\mathbf{x}_{i} - \hat{\mathbf{x}}_i\|^{2},
\end{align*}
where $\hat{\mathbf{x}}_i = D_{\theta'}(E_{\theta}(\tilde{\mathbf{x}}_i))$ denotes the prediction for the $i$ th corrputed vector. 

\section{Proposed Method}\label{sec:proposal}
\begin{figure}[h]
  \centering
  \includegraphics[scale=0.48]{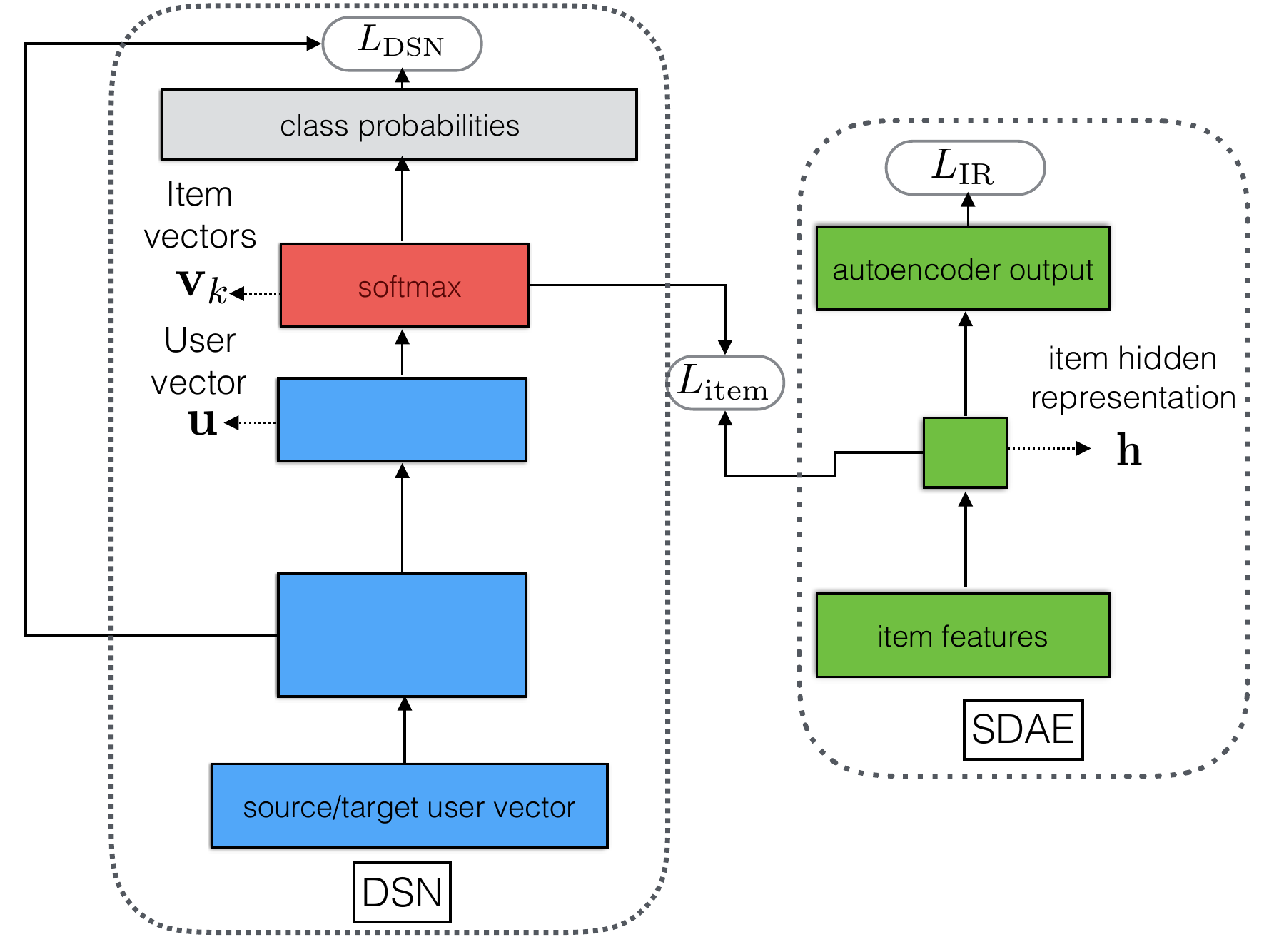}
  \caption{Proposed architecture}
  \label{fig:proposal}
\end{figure}
We first belabor our problem setting. 
To properly suggest items that a user prefer, we learn how likely a user is to choose a certain item from available data. 
Our objective is to learn a classifier that takes a user $\mathbf{x}^{T}$ in the target domain 
as input and predict the probability $p(y = k | \mathbf{x}^{T})$ for $(k = 1, \dots, L)$ that the user would select the item $k$. 

We model this classifier with a DSN model. 
The detail of our model is illustrated in Figure \ref{fig:proposal}. 
The model is a feed-forward neural network that has a DSN 
to obtain an adapted classifier $G$.
For the output layer, we use the softmax layer. That is, the $k$-th element of the output layer is 
\[ \left(G(E_c(\mathbf{x}^{T})\right)_k = p(y=k | \mathbf{x}^{T}) = 
\frac{\exp(\mathbf{v}_k^{\top}\mathbf{u})}{\sum_{k'\in Y}\exp(\mathbf{v}_{k'}^{\top}\mathbf{u})}, \]
where $\mathbf{v}_k$ are softmax weights, and $\mathbf{u}$ is the activations of the previous layer. 
Note that the softmax weight $\mathbf{v}_k$ can be seen as a variable that represents the item $k$, 
and the input vector $\mathbf{u}$ fed into the softmax layer is the representation of a user. 
Given an input vector $\mathbf{x}^T$, we assign a class by $\arg\max_k p(y = k| \mathbf{x}^T)$

As the number of items $L$ is typically large, predictions only with user features 
would be a difficult task. 
In practice, the input data is imbalanced; some labels could be observed just a few times. 
Also, the types of items preferred by users in the target domain might be different from the ones in 
the training data. Therefore, the classifier should be able to predict {\sl new} items not observed in the training data. 
From these observations, we believe that combining item features 
would help predictions. 
In fact, in the video service we used for our experiments, videos that the majority of users watches are short movies of the length of several tens of seconds. Therefore, combining video play times enable the model to identifies which videos are short ones. 
For this purpose, in addition to using a DSN, we also propose combining a denoising autoencoder to incorporate item features. 
We impose the mean squared error between the softmax weight and the hidden representation of 
items calculated by the denoising autoencoder. 

In summary, we train the model so that it minimizes the following loss function,
\begin{align*}
  E = L_{\mathrm{DSN}} +  \lambda_{\mathrm{item}} L_{\mathrm{item}} + \lambda_{\mathrm{IR}} L_{\mathrm{IR}},
\end{align*}
Here, $L_{\mathrm{DSN}}$ is a loss function defined by Eq. \eqref{eq:DSN_loss}.
$L_{\mathrm{item}}$ is the squared loss between the softmax weights and item representations:
\begin{align*}
  L_{\mathrm{item}} = \sum_{i=1}^{N_S} \|\mathbf{v}_{y_i} - \mathbf{h}_i\|^{2}.
\end{align*}
$L_{\mathrm{IR}}$ is the reconstruction loss of the item autoencoder. 
$\lambda_{\mathrm{item}}$ and $\lambda_{\mathrm{IR}}$ are weights that control the interactions of the respective loss terms with the other losses. 

In practical applications, the number of labels typically exceeds thousands, which means an exhaustive evaluation of the softmax function is computationally expensive. To make the computation faster, we use candidate sampling \cite{jean-al-acl2015}. Also, to generate a list of items to recommend at serving time, as mentioned by \citet{Covington:2016}, we only have to perform the neighbourhood search in the dot product space of the learned item features and user features \cite{Liu2004},  
and thus the computation of the softmax function can be avoided.

\section{Experiments}\label{sec:experiment}
We examine the performance of our proposed method in experiments on a pair of real-world datasets. 

\subsection{Dataset Description}
The two datasets consist of the user logs of the following services:
(a) a video on demand service (VIDEO)  
and (b) a news aggregator (NEWS). 
The logs used were collected for three weeks in February in 2017. 
Both datasets contain descriptive attributes of products as follows:
\begin{description}
  \item[VIDEO DATASET] \leavevmode\\
  The implicit feedback of user watches are used. 
  An access to the video page is regarded as a positive example. 
  The dataset contains 48,152 videos of various categories (e.g. film, music). 
  Each video has its title, category, short description, and cast information. 
  To represent a user and an item, we use these features. 
  In addition, we also use the video play time for item representation. 

  \item[NEWS DATASET]\leavevmode\\
  The dataset is comprised of news browsing histories. 
  We treat a user who read an article as a positive example. 
  Each news article has its own title and category, and we use them for feature representation. 
  To make domain adaptation easier, we only use articles of entertainment-related categories, such as music, movie, and celebrity news.
\end{description}
\subsection{Experimental Settings}

We use the first two-week period of the datasets used for training and the rest for evaluation. 
For the source domain (VIDEO), we create a training set of 11,995,769 labeled examples where a label is a video watched by a particular user, 
and the input is a list of previously viewed videos.  
Similarly, for the target domain(NEWS), 
we build a training set of 10,500,000 unlabeled examples that consists of the view histories of news articles. 
For evaluation, we form a test data of size $38,250$ using the logs of users who used both services, which is comprised of pairs of a video label and a news history.

We preprocess the text information of user histories with the TF-IDF scheme.  
We treat a user history  as a document and obtain a vocabulary of 50,000 discriminative words, 
which are chosen according to the tf-idf value. 
In a similar manner, we convert the item's text information. 
An item (video) is treated as a document, and the vocabulary is capped at 20,000 words. 
A video category is transformed to a one-hot vector. 
To extract the features of a video playtime, we convert its hour, minute, and second to one-hot vectors. 
Concatenating these features, we obtain $20,104$ features for items. 

We use the 80 percent of the training data for training and the rest as a validation set. 
Similarly, we sample the 80 percent of the test data. 
We repeat the experiment 5 times by randomly choosing a training-validation and test pair. 

\subsubsection{Evaluation Metric for Generated Candidate Lists}
Given a user history, we generate a list of items sorted according to the relevance to the user. 
We aim to evaluate the goodness of the recommendation. 
To do so, we use the following two metrics:
\begin{itemize}
  \item Recall@$K$
    \begin{align*}
      \mathrm{Recall@}K &= \frac{1}{n}\sum_{i = 1}^n 1_{\left[y_i \in \{\mathrm{top\ }K \mathrm{\ recommended\  items}\}\right]}
    \end{align*}
    $1_{[\cdot]}$ denotes a function which returns $1$ if the statement $[\cdot]$ is true and $0$ otherwise. 
    The index $i$ runs over the test set, and $n$ is the size of the test set, and $y_i$ is the label of the $i$-th example.
  \item nDCG@$K$
    \begin{align*}
      \mathrm{nDCG}@K = \frac{1}{n}\sum_{i = 1}^n \sum_{k=1}^K\frac{1_{\left[y_i = \mathrm{the\ } k\mathrm{th\ suggested\ items}\right]}}{\log(k+1)}
    \end{align*}
\end{itemize}
Unlike rating predictions, in implicit feedback, 
unobserved objects (zero ratings) are uncertain as to whether users have not rated or do not like them.  
Given that, as we only know that positive ratings are true positive, we use recall and nDCG as the evaluation metrics. 
With these metrics, we compare our methods with baseline algorithms. 

\subsubsection{Model Description}
First, we describe the setting for our method. 
We construct a domain separatin network consisted of fully-connected layers with the hidden units: 
\begin{itemize}
  \item the shared/private encoders: $(256-128-128-64)$ \footnote{The left side is the input layer and the right side is the output} ,
  \item the shared decoder: $(128-128-256)$, 
  \item the classifier: $(256-256-256-64)$,
  \item the binary classifier: $(1024-1024)$. 
\end{itemize}
In addition, for item representation, we use a stacked denoising autoencoder with the architecture (20000-256-64-256-20000). 
We denote a DSN with a SDAE by {\bf I-DSN}.   

The exponential linear unit (ELU) \cite{ClevertUH15} is used as the activation function for all layers. The weight and bias parameters for the fully-connected layers are initialized following \cite{HeZR015}. 
To achieve better generalization performance, we apply dropout \cite{dropout} to all fully-connected layers. 
For DSN, the dropout rate is set to $0.75$ for all the encoders and $0.5$ for 
the decoder and the classifier. 
For SDAE, the dropout rate is set to $0.9$ for the input layer and $0.5$ for the rest of layers. 
We also impose the weight decay penalty on the weight parameters with the regularization parameter chosen from $\{10^{-1}, 10^{-2}, 10^{-3}, 10^{-4}\}$. 
The parameters $\alpha$, $\beta$, and $\gamma$ in Eq.\eqref{eq:DSN_loss} are set to $10^{-3}$, $10^{-2}$, $100$, respectively.  
For I-DSN, the $\lambda_{\mathrm{item}}$ and $\lambda_{\mathrm{IR}}$ are fixed at $10^{-2}$ and $100$. 
We use the ADAM optimizer for optimization \citep{DBLP:journals/corr/KingmaB14}. The initial learning rate is fixed at $10^{-3}$. 
Although we aim to perform unsupervised domain adaptaiton, we use a validation set that consists of logs of common users on the same dates as the training data for model selection to investitigate attainable performance of deep domain adaptation.  


\subsubsection{Baseline Description}
In our experiments, we compare our method with the following baselines.
\begin{description}
  \item[Most Popular Items (POP)]\leavevmode\\
  This suggests most popular items in the training data.
  The comparison of our method with this shows how well our method achieves personalization.
  \item[Cross-domain Matrix Factorization (CdMF)]\citep{Iwata:2015}\leavevmode\\
    This is a state-of-the-art collaborative filtering method for cross-domain recommendation. 
    By scraping the same user logs we used for our approach, we obtain a training dataset.
    We eliminte users with history logs fewer than $5$ for the video data and $20$ for the news data. 
    We construct a user-item matrix for each domain, with the value of observed entries $1$ and of unobserved entries $0$. 
    The unobserved entries are randomly sampled. 
    The statistics of the processed data set is given in Table \ref{tab:matrix_statistics}.
    As with our method, we use the $80$ percent of the instances for training and the rest for validation. 

    We initialize the latent vectors with the solution of maximum a posteriori (MAP) estimation as suggested in the paper \cite{Iwata:2015}. 
The dimensionality of latent vectors is set to $10$. 
Although $10$ is small, 
as the training involves the computation of the inverse matrices of this dimensionality for MCMC sampling, it takes very long time even for the dimensionality $10$. 
Also, to make the computation faster, we truncate the iteration at the $700$ sample and discard the first $100$ samples as burn-in.
The hyper parameters $\alpha$, $\beta_0$, and $\nu_0$ are set at the same values as in \cite{Iwata:2015}. 
Because of the time computational complexity, it is hard to optimize these parameters. Therefore we make these parameters fixed. 
  \item[Neural Network (NN)]\leavevmode\\
  To investigate the performance of domain adaptation, we also test the neural network without domain adaptation. 
  We use the same architecture as DSN except for the domain adaptation component by setting $\beta$, $\gamma$ in Eq.\eqref{eq:DSN_loss} to $0$. 
  Parameters for weight decay are chosen from the same parameter grid as our method. 
\end{description}

\begin{table}[h]
  \centering
    \caption{The data statitics for CdMF}
    \label{tab:matrix_statistics}
    \begin{tabular}{r r r}
    \hline
     & News & Movie \\
    \hline
    Instance& $2,258,581$ & $2,073,177$\\
    User& $200,000$ & $200,000$\\
    Item& $7,362$ & $11,648$\\
    \hline
  \end{tabular}
 \end{table}

\subsection{Results of Experiments}
\begin{table*}[h!!]
  \centering
  \caption{Performance comparison of I-DSN, DSN, NN CdMF, POP} 
  \label{tab:comparison_softmax}
  Neural network models are selected according to the valud of the softmax cross entropy loss. 
    \subcaption{NDCG}\label{recall_result}
    \label{tab:comparison_ndcg}
    \begin{tabular}{l l l l l}
      \toprule
      Method & nDCG@$1$ & nDCG@$10$ & nDCG@$50$ & nDCG$100$\\
      \midrule
        I-DSN
        & $0.0260\pm0.0192$ 
        & $0.1671\pm0.0312$ 
        & $0.2593\pm0.0179$ 
        & $0.2668\pm0.0166$\\
        DSN
        & $\mathbf{0.0406\pm0.0212}$
        & $0.1668\pm0.0384$
        & $0.2583\pm0.0230$
        & $0.2655\pm0.0229$\\
        NN
        & $0.0282\pm0.0301$
        & $0.1616\pm0.0279$
        & $0.2473\pm0.0247$
        & $0.2556\pm0.0238$\\
        CdMF
        & $0.0005\pm0.0000$
        & $0.0040\pm0.0000$
        & $0.0135\pm0.0000$
        & $0.0644\pm0.0004$ \\
        POP
        & $0.0398\pm0.0007$
        & $\mathbf{0.2099\pm0.0012}$
        & $\mathbf{0.2790\pm0.0016}$
        & $\mathbf{0.2871\pm0.0010}$\\
      \bottomrule
  \end{tabular}
  \centering
    \subcaption{Recall}

    \begin{tabular}{l l l l l}
      \toprule
      Method & Recall@$1$ & Recall@$10$ & Recall@$50$ & Recall@$100$\\
      \midrule
        I-DSN
        &${0.0260}\pm0.0192$
        &$0.3524\pm0.0641$
        &$\mathbf{0.7487\pm0.0252}$
        &$\mathbf{0.7951\pm0.0281}$\\
        DSN
        &$\mathbf{0.0405\pm0.0212}$
        &$0.3501\pm0.0598$
        &$0.7406\pm0.0145$
        &$0.7842\pm0.0099$\\
        NN
        &$0.02824\pm0.0301$
        &$0.3626\pm0.0281$
        &$0.7292\pm0.0395$
        &$0.7803\pm0.0221$\\
        CdMF
        &$0.0005\pm0.0000$
        &$0.0093\pm0.0001$
        &$0.0551\pm0.0002$
        &$0.3793\pm0.0021$\\
        POP
        &$0.0398\pm0.0007$
        &$\mathbf{0.4472\pm0.0018}$
        &$0.7427\pm0.0015$
        &$0.7916\pm0.0013$\\
      \bottomrule
  \end{tabular}
\end{table*}

\begin{table*}[h!]
  \centering
    \caption[aaaaa]{\protect Performance comparison of I-DSN, DSN, NN, POP.}
    \label{tab:comparison_ndcg2}
    Neural network models are selected according to the value of nDCG@$100$
    \begin{tabular}{l l l l l}
      \toprule
      Method & nDCG@$1$ & nDCG@$10$ & nDCG@$50$ & nDCG$100$\\
      \midrule
        I-DSN
        &$0.0440\pm0.0256$
        &$0.1881\pm0.0282$
        &$0.2693\pm0.0211$
        &$0.2785\pm0.0192$\\
        DSN
        &$\mathbf{0.0618\pm0.0212}$
        &$\mathbf{0.2133\pm0.0154}$
        &$\mathbf{0.2873\pm0.0151}$
        &$\mathbf{0.2945\pm0.0153}$\\
        NN
        &$0.0415\pm0.0211$
        &$0.1938\pm0.0131$
        &$0.2735\pm0.0102$
        &$0.2797\pm0.0107$\\
        POP
        & $0.0398\pm0.0007$
        & $0.2099\pm0.0012$
        & $0.2790\pm0.0016$
        & $0.2871\pm0.0010$\\
      \bottomrule
  \end{tabular}
\end{table*}

In Table \ref{tab:comparison_softmax}, we show the performance comparison in terms of the two evaluation metrics. 
For I-DSN, 
at test time, for unobserved items, we can initialize the softmax weights with the hidden representation of the items returned by the autoencoder. 
Although we tried this approach but it did not yield better results compared to when the weights were not initialized. Therefore, we report the results of I-DSN with the weights of unobserved items not initialized with the hidden representations.

In both results, CdMF underperforms the other methods. 
As mentioned in \cite{Hu2008}, it can be considered that CdMF cannot process the implicit feedback and accurately express the popularity structure in the dataset.  

We can see that the domain adaptation approaches outperforms NN in terms of both metrics. 
In terms of recall,  I-DSN shows the better average performance compared to POP when the length of suggested lists $K$ is $50$ and $100$, while DSN cannot beat the results of POP. 
However, in regards to nDCG, both domain adaptation methods underperform POP. 
This results suggests that optimizing the model in terms of the softmax cross entropy loss could produce higher recall outputs but not ones of good ranking quality. 

Based on the above results, we aim to produce a model that has good ranking performance. 
To do so, we calculate nDCG@$100$ on the validation set and choose the model by changing the weight decay parameters and the iteration step. 
The result is shown in Table \ref{tab:comparison_ndcg}.
While all three methods show improvement in their result, 
DSN shows the best average performance compared to the other methods. 
Thus, by aligning the objective loss with the evaluation metric, DSN can improve its recommendation performance.

In summary, it has been shown that the performance of a simple neural network was improved by using domain adaptation even in cross-domain recommendation.  In addition, our approaches outperformed a state-of-the-art collaborative filtering method. 
Although the domain adaptation approach overcame other machine learning methods, however, deep domain adaptation sometimes underperformed a simple popularity-based method. 
Specifically, our method underperformed the popularity-based method particularly when the length of suggested lists is short. 
This indicates that for services that can suggest a small number of items, only using our approach may not be effective. 
However, our model can produce a better candidate set with higher recall when the length of suggested items is longer. 
Also, predicting items from it using other (heuristic) ranking procedures, e.g. combining the result of popularity-based ranking, could improve the result. 
Furthermore, the application of domain adaptation showed actual improvement when the loss function is aligned with the evaluation metric. Therefore, we conclude that we obtained positive results from these experiments.

\section{Conclusions and Discussions}\label{sec:conclusion}
In this paper, we studied a problem of cross-domain recommendation in which we cannot expect common users or items and presented a new deep learning approach using domain adaptation. 
We investigated the performance of our approach in cross domain recommendations by conducting experiments on a pair of two real-world datasets. We confirmed that our approach actually improved state-of-the-art in the literature. 
Although the improvement was not overwhelming, we showed that by treating the problem as extreme classification, domain adaptation was applicable and 
obtained promising results that encourages the further extension of our model. 
As this is the first attempt in the literature, based on these results, we would like to keep improving in future work. 



Specifically, 
We envision that further improvement can be achieved in the following three ways. 
First, item feature extraction. 
Combining item representation with a stacked denoising autoencoder showed the improvement in recall. 
However, initializing softmax weights with item features did not produce a better result. 
This implies that the extracted features from autoencoder only contribute to predicting labels observed in the training data. 
We hypothesize that this could be because only using the text information and the length of videos does not make each video distinct from others. 
To achieve better generalization performance, we should include other types of information about videos. 
For example, if a certain program is episodic, then implementing such information may help the prediction 
because we can deduplicate the same type of videos from the suggested list by treating the videos as one program. 

Second, as with item feature extraction, we need improve the training of the classifier so that 
it can work for classes with few examples. 
In terms of recall, there was not overwhelming difference between the popularity method and our approach. 
This indicates that to have better performance, our classifier should be able to better predict rare (not just popular) items. 
Although dealing with class imbalance is a persisting issue, this could be realized, for example, by combining techniques of few-shot learning \cite{Vinyals2016, KaiserNRB17,finn2017a}. 

Third, as optimizing the model with nDCG saw a large improvement, we also suggest replacing the softmax loss with another ranking loss such as nDCG \cite{NIPS2009_3758}.

\bibliographystyle{ACM-Reference-Format}
\bibliography{ms} 

\end{document}